\documentclass{article}

\usepackage{PRIMEarxiv}

\usepackage[utf8]{inputenc} % allow utf-8 input
\usepackage[T1]{fontenc}    % use 8-bit T1 fonts
\usepackage{hyperref}       % hyperlinks
\usepackage{url}            % simple URL typesetting
\usepackage{booktabs}       % professional-quality tables
\usepackage{amsfonts}       % blackboard math symbols
\usepackage{nicefrac}       % compact symbols for 1/2, etc.
\usepackage{microtype}      % microtypography
\usepackage{lipsum}
\usepackage{fancyhdr}       % header
\usepackage{graphicx}       % graphics
\graphicspath{{media/}}     % organize your images and other figures under media/ folder
\usepackage{multirow}%
\usepackage{multicol}
\usepackage{amsmath,amsfonts}%
\usepackage{amsthm}%
\usepackage{mathrsfs}%
\usepackage[title]{appendix}%
\usepackage{xcolor}%
\usepackage{textcomp}%
\usepackage{manyfoot}%
\usepackage{booktabs}%
\usepackage{listings}%
\usepackage{subfig}
\graphicspath{{./}}
\usepackage{float}
\usepackage{soul}
\title{Decorrelation-based Self-Supervised Visual Representation Learning for Writer
Identification
}

\author{
  Arkadip Maitra \\
  Department of Computer Science \\
  Ramakrishna Mission Vivekananda Educational and Research Institute, Belur \\
  Belur, India\\
  \texttt{arkadipmaitra@gmail.com} \\
   \And
  Shree Mitra \\
  Department of Computer Science and Engineering \\
  Indian Institute of Information Technology, Guwahati \\
  Guwahati, India\\
  \texttt{shree.mitra23m@iiitg.ac.in} \\
  \And
  Siladittya Manna \\
  CVPR Unit \\
  Indian Statistical Institute, Kolkata \\
  Kolkata, India\\
  \texttt{siladittya\_r@isical.ac.in} \\
  \And
  Saumik Bhattacharya \\
  Dept. of EECE \\
  Indian Institute of Technology, Kharagpur \\
  Kharagpur, India\\
  \texttt{saumik@ece.iitkgp.ac.in} \\
  \And
  Umapada Pal \\
  CVPR Unit \\
  Indian Statistical Institute, Kolkata \\
  Kolkata, India\\
  \texttt{umapada@isical.ac.in} \\
}

\begin{document}
\maketitle

\begin{abstract}
  Self-supervised learning has developed rapidly over the last decade and has been applied in many areas of computer vision. Decorrelation-based self-supervised pretraining has shown great promise among non-contrastive algorithms, yielding performance at par with supervised and contrastive self-supervised baselines. In this work, we explore the decorrelation-based paradigm of self-supervised learning and apply the same to learning disentangled stroke features for writer identification. Here we propose a modified formulation of the decorrelation-based framework named SWIS which was proposed for signature verification by standardizing the features along each dimension on top of the existing framework. We show that the proposed framework outperforms the contemporary self-supervised learning framework on the writer identification benchmark and also outperforms several supervised methods as well. To the best of our knowledge, this work is the first of its kind to apply self-supervised learning for learning representations for writer verification tasks.
\end{abstract}

% keywords can be removed

\section{Introduction}
Writer identification and signature verification are essential steps for identity verification of specific documents like forms, bank cheques, or even the writer of any old book or text. Hence, tasks like Writer identification and signature verification are very important and sensitive tasks in the domain of computer vision and pattern recognition domain. Writer identification is related to finding the author of a given document based on the learned features or descriptors. Depending on the contents of the query and reference document, the writer identification technique can be divided into text-dependent or text-independent approaches. The text-dependent approach requires the same text to be written while the text-independent approach does not require any particular text. However, the text-independent approach is more complicated than the text-dependent approach because it must extract writer-specific features regardless of signatures, specified letters, or symbols to be compared \cite{NGUYEN2019104}. In the work, we focus on solving text-independent writer identification tasks. %The author needs to reproduce a sample of the given text based on which the identity is established.

Most of the recent works in the field of writer identification \cite{zhang2023cmtco, he2021grrnn} have used supervised techniques. With the advent of self-supervised learning (SSL) techniques \cite{chen2020simclr, he2019momentum, Zbontar2021BarlowTS}, applications of SSL in the computer vision domain have rapidly grown over the last few years. The primary objective of self-supervised methods is to learn representations from unlabeled data. While SSL has gained much attention from the computer vision community, the application of SSL in document image analysis has not gathered momentum. While there are applications like document layout segmentation \cite{maity2023selfdocseg, li2021selfdoc}, instances are not plenty. Recently, \cite{Manna2022SWISSR, chattopadhyay2022surds, luo2022siman} are some of the groundbreaking literature in document or text image-related works.

\begin{figure}[t]
    \centering
    \begin{tabular}{cc}
        \subfloat{\includegraphics[width = 0.45\linewidth]{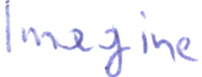}} & 
        \subfloat{\includegraphics[width = 0.45\linewidth]{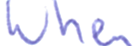}}\\
    \subfloat{\includegraphics[width = 0.45\linewidth]{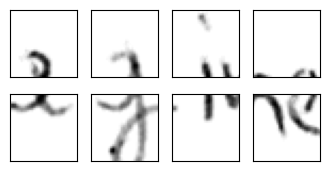}} & 
        \subfloat{\includegraphics[width = 0.45\linewidth]{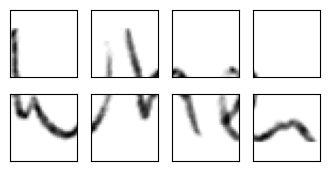}}\\
    \end{tabular}
    \caption{Samples images from the CVL dataset show the sparse nature of the handwritten text dataset images}
    \label{fig:example}
\end{figure}

While SSL is one of the best tools for representation learning, it suffers from its own issues. One such issue is the collapse of representations. There are primarily two types of collapse, dimensional collapse and singularity or complete collapse \cite{li2022dimcol}. Dimensional collapse means that the representation embeddings occupy a lower-dimensional subspace than their actual number of dimensions. It blocks the learning of representations by diminishing the separability of the high-dimensional latent space, as an exponential number of samples are better separable in a high-dimensional space than in a low-dimensional space (by Cover's Theorem \cite{cover1965coverstheorem}). 
Dimensional collapse is mainly observed in contrastive learning frameworks. Some of the methods used to prevent such collapse are the use of non-linear projector \cite{chen2020simclr}, momentum encoding \cite{he2019momentum}, stop-gradient \cite{simsiam}, etc. Another remedy to such collapse is using decorrelation \cite{hua2021featdecorr}. The primary idea behind decorrelation is to train the model such that each output feature dimension is uncorrelated with each other. As a result of that, the information obtained from each feature dimension will not be linearly correlated with each other. We show in Sec. \ref{subsec:motivation}, under the assumption that the dimensions follow a joint multi-variate distribution, the feature dimensions are also independent. This assumption is also made in \cite{Zbontar2021BarlowTS}, but the authors have not mentioned the reason for such an assumption in detail when they are only focusing on the decorrelation operation.

Another important point in writer identification tasks is the sparse nature of the handwritten text images. As can be seen from Fig. \ref{fig:example}, each word image mostly contains white pixels, with little stroke information in each image. Such sparse data is not suitable for contrastive learning algorithms, as it is prone to collapse. The correlation between such images can be minimized but can never be zero. Even when divided into patches (Sec. \ref{subsec:featextrac}), the correlation between the constituent strokes will be considerable, considering the number of white pixels in each patch.

The self-supervised technique proposed in this work follows a decorrelation-based objective, similar to \cite{Manna2022SWISSR, Zbontar2021BarlowTS}. In addition to decorrelating the output dimensions, we also assume that the output dimensions belong to a joint multivariate distribution, to ensure the disentanglement of the output dimensions. Following this argument, we expect the model to learn fundamental stroke features, such that each word can be represented in terms of its constituent disentangled stroke features.

The primary contributions of our work are as follows:
\begin{itemize}
    \item To the best of our knowledge, this work is the first to propose a self-supervised learning framework for learning representation for writer identification tasks.
    \item We propose a novel SSL framework based on decorrelation for learning disentangled stroke representations.
    \item We mathematically explain in detail the concept of disentanglement in relation to decorrelation and independence.
    \item We statistically analyze the correlation between the constituent patches to verify our claims.
    \item The proposed framework outperforms the contemporary SSL framework on the writer identification benchmark and also outperforms several supervised methods as well.
\end{itemize}

\section{Related Works}
This work builds on prior work in several domains: handwritten text recognition, signature recognition, self-supervised learning, and recent self-supervised learning for writer identification and signature verification.
\subsection{Writer Identification}
In literature, Writer Identification and Verification has been an effective domain of research and development in the field of document image processing for more than two decades \cite{halder2014iibc}. A writer identification approach based on a bag of words with OBI attributes was proposed by Durou et al. \cite{DUROU2019BOI}. The authors have proposed to deploy a combination of both methods, one using Oriented Basic Image features and the concept of graphemes codebook and another one to reduce the resulting high dimensionality of the feature vector a Kernel Principal Component Analysis has been used. % The authors have achieved 96\% accuracy on the IAM dataset for English handwriting and the ICFHR 2012 dataset for Arabic handwriting. 
A method for content-independent writer identification on Bangla text images was proposed by Halder et al. \cite{halder2018document}. In this approach, the authors have used local handwriting-based attributes as features, a multi-layer perception, and a simple logistic classifier for classification purposes. The authors have tested this method on an unconstrained or in other words an informal handwritten Bangla database of 1383 documents from 190 writers. % This method surpasses the state-of-the-art methods by approximately 3\% (top-3 and top-4 choices) and is 27 times faster than the conventional segmentation method. 
A method for a Chinese character-level writer identification system using path signature features and deep convolutional neural networks was proposed by Yang et al. \cite{yang2015dropstroke}. The authors have proposed a data augmentation approach, DropStroke to enrich personal handwriting. The authors have conducted an experiment using this method on online handwritten Chinese characters from the CASIA-OLHWDB1.0 data set \cite{wang2009casia}, which consists of 3866 classes from 420 writers. The authors have got encouraging results by using this method. A method that uses a convolutional neural network (CNN) based deep learning method for writer identification was proposed by Rehman et al. \cite{rehman2019avf}. % The authors have got an accuracy of 92.78\% on English, 92.20\% in Arabic, and 88.11\% on the combination of Arabic and English, respectively. 
A novel system for writer identification based on the pre-segmented characters of the Devanagari script was proposed by \cite{dargan2020svm}. The authors have used K-NN and SVM classifiers for the classification task. %and they have got an identification accuracy of 91.53\%. 
An approach for writer identification from handwritten Bangla characters and numerals was proposed by Adak et al. \cite{adak2015numerals}. In this paper, the authors have used an SVM classifier for the classification task. For experimental analysis, a database containing 2,12,300 isolated Bengali characters and numerals is generated with the help of 100 writers. The writers have got promising results using this approach. A method of offline Bengali writer verification PDF-CNN and Siamese Net was proposed by Adak et al. \cite{adak2018pdf}. In this paper the authors have extracted some important handwritten features from the handwritten text and then input the probability distribution function (PDF) of that feature into the convolutional neural network (CNN) after that they got a hybridized feature which was further fed into the  Siamese neural network for writer verification. In this paper, the authors have used two handcrafted features, one is textural-based features and another one is allographic-based features. The networks used to create the hybrid features are named Textural PDF-CNN and allographic PDF-CNN respectively. For the writer identification, they used an MLP and in another case a Siamese Net after the CNN. They have designed 4 types of architecture: Textural CNN-MLP, Allographic CNN-MLP, Textural CNN-Siamese, and Allographic CNN-Siamese. % and they have got an accuracy of 96.49\%, 95.92\%, 97.64\%, and 96.47\% respectively. 
Another method for text-independent writer identification using convolutional neural network(CNN) was proposed by Nguyen et al. \cite{NGUYEN2019104}. % where the authors have got 99.97\% accuracy to classify 100 writers by 200 characters for handwritten Japanese and 91.81\% accuracy to classify 900 writers by one text page for handwritten English.
\subsection{Self-Supervised Learning}
Early self-supervised methods often designed self-supervised tasks, like pretext tasks \cite{gidaris2018rotnet} created a pretext task by rotating the original image by specific angles and then let the network learn semantic representations with the rotation prediction task. Another pretext task proposed by \cite{doersch2015conpred} extracted random pairs of patches from each image and trained the network to predict the position of the second patch relative to that of the first. Recent self-supervised techniques using contrastive learning have shown huge improvement over their manually-designed pretext task-based counterparts. The paper \cite{he2019momentum} used InfoNCE \cite{oord2018cpc} contrastive loss between representations of a pair of differently augmented views of the same image obtained from an online encoder and a momentum updated encoder. Another landmark work on contrastive learning named SimCLR \cite{chen2020simclr} showed the varying effects of different augmentations on the InfoNCE-based contrastive learning framework besides proposing an InfoNCE loss-based self-supervised learning framework. In contrast to MoCo \cite{he2019momentum}, the SimCLR framework obtained state-of-the-art results without using a feature memory bank. This paper also showed massive improvement in encoding quality when batch size was increased and network size was increased. However, as studied and explained in \cite{hua2021featdecorr} and \cite{li2022dimcol}, contrastive learning algorithms are prone to the dimensional collapse of representations. Although several remedies like the use of a projector, momentum encoding \cite{he2019momentum} or stop gradient \cite{simsiam} is used, those additional architectural or training methodologies are not clearly understood \cite{li2022dimcol}. 

A separate line of thought led to the advent of non-contrastive methods like BYOL \cite{grill2020byol}, Barlow twins \cite{Zbontar2021BarlowTS}, etc. BYOL and Barlow Twins both use a projector in addition to the base encoder. However, as mentioned in \cite{hua2021featdecorr}, decorrelation-based methods are able to prevent dimensional collapse. BYOL minimizes the L2 distance between a feature vector obtained from a sample of the positive pairs and its predicted version from the other sample in the positive pair. The framework proposed in Barlow Twins \cite{Zbontar2021BarlowTS} aims at decorrelating the feature dimension by minimizing the cross-correlation between any two feature dimensions to zero and the auto-correlation values to one.

The application of SSL in handwritten text-related tasks is very limited. Some of the initial works in this domain are \cite{Bhunia2021VectorizationAR, chattopadhyay2022surds, Manna2022SWISSR}. The work proposed in  \cite{Bhunia2021VectorizationAR} is based on an online signature dataset and utilizes inter-conversion between vectorization and rasterization data as the supervision signal. \cite{chattopadhyay2022surds} takes a reconstruction-based approach to learning stroke information from signature images. The work proposed in \cite{Manna2022SWISSR} is also based on signature verification tasks. \cite{Manna2022SWISSR} also takes a decorrelation-based approach to learning stroke representations from L2-normalized feature dimensions. 

\begin{figure*}[!ht]
    \centering
    \includegraphics[width = 0.99\linewidth]{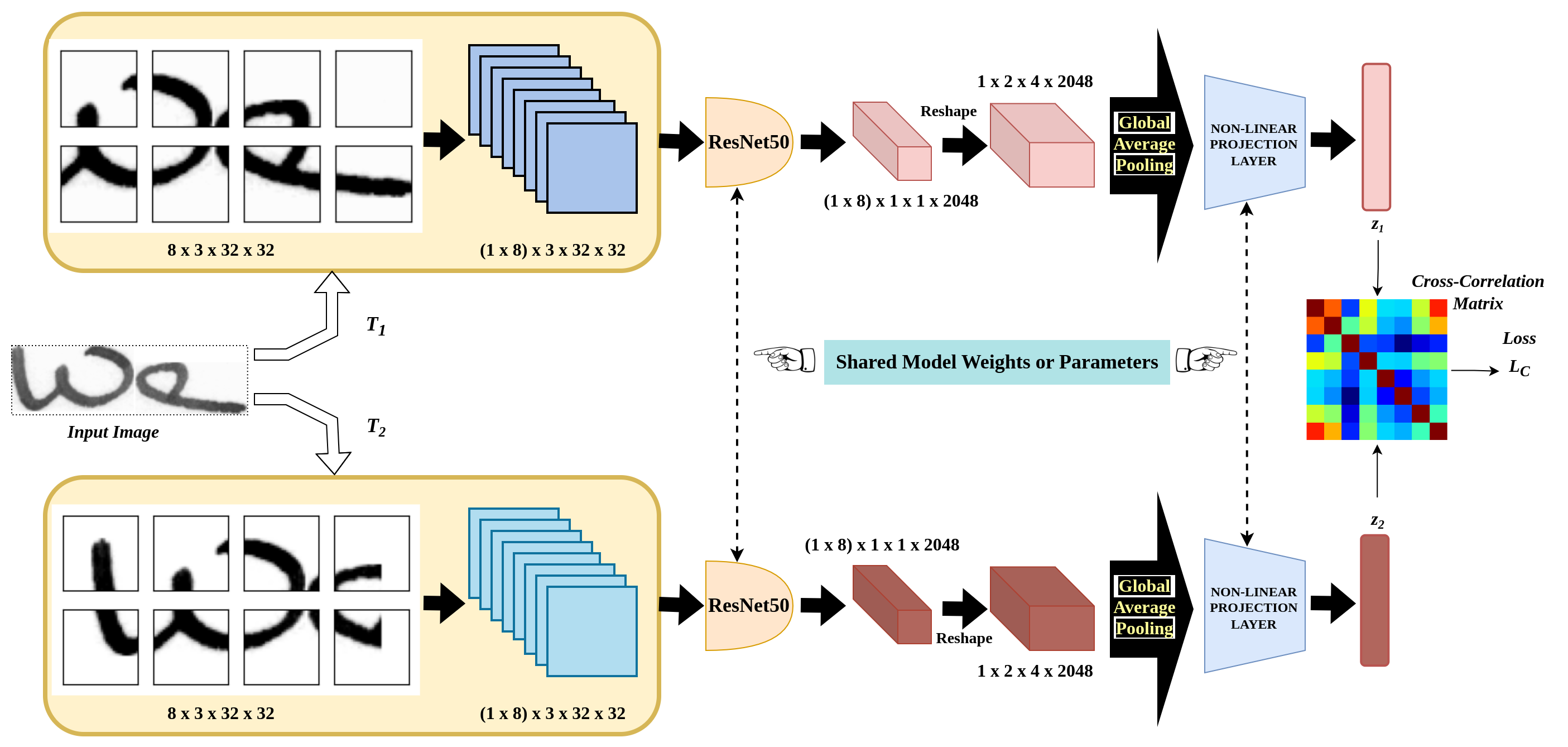}
    \caption{Figure showing the proposed framework. A single image is augmented to form a positive pair, which is fed to the weight-shared base encoder for feature extraction. After reshaping the output from the encoder, the feature vectors are passed through the projector. The final feature vector is passed to the loss function for loss calculation and subsequent optimization. $T_1$ and $T_2$ are the two different augmentations applied on the input image to obtain the positive pair.}
    \label{fig:model}
\end{figure*}

Thus, the framework proposed in this work is the first of its kind to apply self-supervised representation learning in the domain of text-independent writer identification, to the best of our knowledge.

\section{Methodology}

\subsection{Feature Extraction}
\label{subsec:featextrac}
In handwritten images, it is essential to capture the stroke information from the different authors as well as to learn the variations in the handwriting of the same individual. Any other distortions would have influenced the handwriting strokes which would have been counter-productive to the pretraining goal.

For word-level writer identification, to capture enough information each word image is transformed to a shape of 64x128 either by center cropping or by white padding the word image. The input images are then divided into patches of dimensions 32x32 and fed as input to the encoder. From each image, we get 8 patches. We stack the patches along the batch dimension to form $(N \times 8) \times 3 \times 32 \times 32$, where $N$ is the total number of images in a batch. After the features are extracted from the encoder and projector, we reshape the output feature map to shape $N \times 2048 \times 8$. Before taking a global average pooling over the spatial dimensions, the output is reshaped to $N \times 2048 \times 2 \times 4$. The pooled feature vector of shape $N \times 2048$ is finally used in the optimization of the model parameters.

The base encoder used in our proposed framework is a ResNet50 \cite{he2016resnet} model, except for the last classification layer. Following \cite{Manna2022SWISSR, Zbontar2021BarlowTS}, we attach a projector to the end of the encoder. The projector is a 2-layered multi-layered perception model, which takes an input with 2048 channels and outputs a feature vector with 2048 channels. To account for the small input dimensions, the ResNet50 architecture was modified according to the changes mentioned in \cite{chen2020simclr}. The first convolutional layer was replaced with a convolutional layer with $3 \times 3$ kernel and 64 channels. The subsequent max-pooling layer was also removed. 

Before proceeding further, it is important to define the term \textit{positive pair}. A positive pair is composed of two differently augmented versions of the same image. The final loss is calculated between two sets of samples obtained from a batch of positive pairs.

\subsection{Motivation of the Proposed Loss Function}
\label{subsec:motivation}
% \hl{Review This subsection}

In our work, the objective is to learn decorrelated representations such that each word can be learned as a spatial combination of several fundamental strokes. This also necessitates that the fundamental stroke representations are distinguishable from each other to prevent a collapse of representations. In \cite{Manna2022SWISSR} it was also assumed that each output feature dimension will learn representations that are representative of each fundamental stroke. From \cite{hua2021featdecorr}, we already know that the decorrelation mechanism is an important tool to prevent the aforesaid collapse. To ensure the aforementioned objective, it is important that the point in the latent space where a stroke is mapped is farther away from any other dissimilar stroke.

In this work, we take a different approach to decorrelation. In addition to normalizing each feature dimension to a unit-normalized vector, we also whiten each dimension. If we consider the feature vector for each sample to be a random variable $\mathcal{X} = \{x_1, \hdots, x_D\}$, such that $\mathcal{X} \in \mathbb{R}^D$. We can say, that for the loss function proposed in \cite{Manna2022SWISSR}, $x_i \sim \mathcal{N}(0,1)$ and $\mathcal{X} \sim \mathcal{N}(0, \Sigma)$, where $\Sigma$ is the covariance matrix. Each feature dimension is treated as a normal distribution and the covariance matrix is converted to a diagonal matrix.

\paragraph{Understanding the prevention of dimensional collapse of representations} We adopt a different normalization strategy than in SWIS \cite{Manna2022SWISSR}, where $l2$-normalizing the along the batch dimension converts each feature dimension to a uniform distribution on a unit hypersphere $\mathbb{S}^{N-1}$ \cite{muller1959, marsaglia1972}. But the dimensionality of the hypersphere $\mathbb{S}^{N-1}$ is dependent on the batch dimension. If each dimension of the random variable $x_i$ of $\mathcal{X}$ is mapped to the surface of a hypersphere $\mathbb{S}^{N-1}$, where $N$ is the number of samples in a batch and is distributed according to a uniform distribution \cite{muller1959, marsaglia1972}, then the collapse of the representations is already inhibited by constraining the samples to be uniformly distributed. Furthermore, a dependence on the number of samples in a batch also introduces variability on a hyper-parameter which is not an intrinsic property of the base model.

\paragraph{Understanding Decorrelation}
Now, it is important to understand that, the value that $x_i$ takes for each sample along the dimension $i$ is drawn from a uniform distribution defined on a unit hypersphere $\mathbb{S}^{N-1}$. It is common knowledge that, two variables that are independent are always uncorrelated, but the reverse is not true. %While these statements are standard, we still prove them in the manuscript for brevity.

However, if the variables are jointly normally distributed, then the reverse is also true, that is, if we have two uncorrelated bivariate normally distributed variables, we can say that they are independent.

 If two dimensions $d_i$ and $d_j$ are bivariate normal and uncorrelated, then they are independent. For disentanglement to hold, the variables need to be both independent and uncorrelated. To this end, we first standardize the variables by subtracting the mean and the unbiased standard deviation $\left(\hat{\sigma} = \sqrt{\frac{1}{N-1} \sum_{i=1}^{N} (x_i - \bar{x})^2}\right)$, which is the best linear unbiased estimator (BLUE) for any distribution. 

However, in our case, to disentangle any two dimensions $d_i$ and $d_j$, we need to first assume that these two dimensions are normally distributed. Under the aforesaid assumption, we standardize along each $d_i \; \forall i \in [1, D]$, which centers the mean to $0$ and sets the variance to $1$. %If the number of samples is large enough, then according to the Central Limit Theorem we get, $x_i \sim \mathcal{N}(0,1)$. 
Since, we do not have knowledge about the population distribution, we assume it to follow a normal distribution. Considering the above observations, to ensure disentanglement, we need to decorrelate any two dimensions $d_i$ and $d_j$.

Take note that the conclusion reached in the aforesaid statements is not the same as the ones stated in \cite{Manna2022SWISSR}. In \cite{Manna2022SWISSR}, although the authors assumed that the random variable $x_i$ representing each dimension $d_i$ is normally distributed, the authors did not standardize each dimension, even though the objective was devised to reduce the variance of each dimension to $1$. The standardization step also makes our work different from \cite{Zbontar2021BarlowTS}, which also adopted a similar decorrelation based approach.

\subsection{Proposed Loss function}
After the features from the sample in a positive pair are extracted from the encoder and projector, the features are given as input to the loss function. The loss function used for learning representation in the pre-training phase is as follows,
\begin{equation}
    L_c = \frac{1}{N}\sum_{i = 1}^D(\sum_{j = 1, j \neq i}^D(\sum_{k = 1}^N z_k^i.z'^j_k)^2 + (\sum_{k = 1}^Nz_k^i.z'^i_k-1)^2)
\end{equation}
where $z^i_k$  is a scalar value at $i$-th dimension of the $k$-th centered and normalized feature vector $z_k$.Thus, the preprocessing steps before feeding the feature vector $z^i_k$ to the loss function are as follows:

\paragraph{\textbf{Step 1} :}
\begin{equation}
    \overline{z_k^i} = \frac{\widetilde{z}^i_k}{\sqrt{\sum_{k = 1}^N(\widetilde{z}^i_k)^2}} \forall i \in [1, D]
\end{equation}

where $\widetilde{z}^i_k$ is the value of the $k$-th sample at the $i$-th dimension.
\paragraph{\textbf{Step 2} :}
\begin{equation}
    z^i_k = \frac{\overline{z^i_k} - \mu_{z_k}}{\sqrt{\frac{1}{N}\sum_{k = 1}^N(\overline{z^i_k} - \mu_{z_k})^2}} \hspace{0.5 in } \; \text{where} \; \mu_{z_k} = \frac{1}{N}\sum_{k = 1}^N\overline{z_k^i}
\end{equation}
It is to be noted that $z^i_k$ and $z'^i_k$ are obtained from each element of a positive pair. 

By minimizing $L_c$, we disentangle the output feature dimensions and enable the model to learn to disentangle stroke features. In Table \ref{table:5} and \ref{table:6}, we can clearly see that the proposed method improves performance over the current state-of-the-art self-supervised framework SWIS on downstream writer identification tasks.

\subsection{Downstream Task}
\label{subsec:downstreamtask}
The downstream task in this work is a writer identification task. Given a handwritten word image, the model is trained to predict the writer who has written it. The downstream task is trained in two ways: 1) word level and 2) page level. In the word level model, the model is trained to predict the writer for each word image. Whereas in page level model, the model is trained to predict the writer from all the word-level images contained in a page written by a writer. It is to note that, the downstream training was conducted only for the word-level images. For the page-level evaluation, no training was conducted. The model trained on word-level ground truth data was used to evaluate the page-level performance of the proposed model. To do so, we predicted the writer for each word-level image and did majority voting to find the final predicted writer id.

\section{Implementation Details}

\subsection{Datasets}
In this work, to perform the writer identification task, we used three datasets namely CVL\cite{kleber_florian_2018_1492267}, IAM\cite{Marti2002} and Firemaker\cite{Firemaker}. For the training and testing splits used in this work for each of the datasets, we follow the splits used in the work FragNet\cite{he2020fragnet}. The description of each of the datasets is given below.
\subsubsection{\textbf{IAM}}
The dataset contains 13,353 labeled text lines of diverse content written in English, with an average of 14 text lines per writer, and is often used for handwriting authentication. It contains 1,539 forms written by 657 different authors, with specific details like author identity and ground truth text. If a writer submits more than one page, we randomly choose one for testing while using the additional pages for training. We randomly divide text lines into training and testing sets for authors who only contribute one page. This collection includes the word image bounding boxes. % In order to create the training set, we collect word images from the training pages, while word images from the testing pages are used for testing.

\subsubsection{\textbf{CVL}}
This dataset is a bilingual collection of handwritten texts written by 310 distinct authors in both English and German. The database contains seven distinct handwritten texts, each with an RGB both a full-color and cropped version. There were 310 writers who contributed to the dataset, 27 of whom wrote 7 texts, and 283 writers overall making five texts. In this work, we have used the first three texts from each writer for training and the rest of the texts for each writer are used for testing, and the word images are also available for this dataset.

\subsubsection{\textbf{Firemaker}}
This dataset includes 1000 images of scanned handwritten text in grayscale at a resolution of 300 dpi, each including four pages of handwritten text from each of the 250 Dutch writers. In this work, text images from the first page are used for training, and text images from the fourth page are used for testing.

\subsection{Pre-training Experimentation}
For the pre-training phase, we used ResNet50 \cite{he2016resnet} as the base encoder. The model parameters were optimized using Adam optimizer. The learning rate was set to $10^{-3}$. In addition to that, we decayed the learning rate according to a cosine annealing scheduler after a warmup phase of 10 epochs. We trained the model for 500 epochs on the IAM, Firemaker, and CVL datasets.

It is to be noted that the input images were reshaped to spatial dimensions $64 \times 128$. We used a number of augmentations, including color jittering, affine translation, and random cropping to $64 \times 128$, to make sure the pre-trained models learned robust and generalized features. After augmentation, the images were normalized to the range $[-1.0, +1.0]$. We cropped the images so that the input to the encoder comprised a tightly bound handwritten image because not all of the images in the datasets contain correctly cropped signature images. To accomplish this goal, we first used Otsu’s thresholding \cite{4310076}, then we looked for the bounding box with the smallest area that contained all non-zero pixels near the image’s center of mass. Following this preprocessing stage, the images were cropped into non-overlapping patches, each with spatial dimensions $32\times32$-pixel patches, giving us a total of 8 patches from each image. Since ResNet50 only takes 4D input, we arranged the patches from all the images in a batch with batch size $N$ along the batch dimension, giving us an input of shape $(N \times 8) \times 3 \times 32 \times 32$.

\subsection{Downstream Experimentation}
\label{subsec:downstreamexpt}
The downstream training is done so that the pre-trained encoder can be used to classify handwritten images. A linear layer replaces the projection layer with the number of classes as the output. The downstream model is then trained using an Adam optimizer for 500 epochs, with a learning rate of $10^{-4}$. The model parameters were optimized by minimizing the categorical cross-entropy loss.

\section{Experimental Results}
% \subsection{Comparison with State-of-the-Art Supervised Methods}

In this section, we present the results obtained by the proposed framework on the 3 benchmark datasets, IAM, CVL, and Firemaker. We use two levels of evaluation, namely, word level and page level. The procedures for word-level and page-level evaluation are detailed in Sec. \ref{subsec:downstreamtask}.

\begin{table}[!ht]
    \centering
    \caption{Comparison results of the proposed self-supervised framework with the contemporary supervised learning algorithms on Word Level performance in terms of accuracy (\%). }
    \begin{tabular}{cccc}
    \toprule
    \multirow{2}{*}{Methods}& \multicolumn{3}{c}{Datasets}\\ %\cline{2-4}
    & IAM & CVL & Firemaker\\
    \midrule
    \multicolumn{4}{c}{Supervised} \\
    \midrule
    Vert. GR-RNN \cite{he2021grrnn}  & 85.9 & 92.6 & 76.5 \\ 
    Hor. GR-RNN \cite{he2021grrnn}   &  86.1 & 92.4 & 76.9 \\ 
    MSRF-Net \cite{sri2022msrf} &  84.6 & 91.4 & 71.2 \\ 
    WordImageNet \cite{he2020fragnet}& 81.8 & 88.6 & 67.9 \\ 
    FragNet-16 \cite{he2020fragnet}& 79.8 & 89.0 & 59.6 \\ 
    FragNet-32 \cite{he2020fragnet}&  83.6 & 89.0 & 65.0 \\ 
    FragNet-64 \cite{he2020fragnet}& 85.1 & 90.2 & 69.0 \\
    \midrule 
    \multicolumn{4}{c}{Self-Supervised}\\
    \midrule
    SWIS  \cite{Manna2022SWISSR}& 83.91 & 92.76 & 73.01\\ 
    Ours & \textbf{84.8} & \textbf{93.32} & \textbf{74.24} \\
    \bottomrule
    \end{tabular}
    \label {table:5}
\end{table}

\begin{table}[!ht]
    \centering
    \caption{Comparison results of the proposed self-supervised framework with the contemporary supervised learning algorithms on Page Level performance in terms of accuracy (\%). }
    \begin{tabular}{cccc}
    \toprule
    \multirow{2}{*}{Methods}& \multicolumn{3}{c}{Datasets}\\ %\cline{2-4}
    & IAM & CVL & Firemaker\\
    \midrule
    \multicolumn{4}{c}{Supervised} \\
    \midrule
    Vert. GR-RNN \cite{he2021grrnn} &  96.4 & 99.3 & 98.8 \\
    Hor. GR-RNN \cite{he2021grrnn}  &  96.4 & 99.3 & 98.8 \\
    MSRF-Net \cite{sri2022msrf}&  94.8 & 99.4 & 97.2 \\
    WordImageNet \cite{he2020fragnet}&  95.8 & 98.8 & 97.6 \\
    FragNet-16 \cite{he2020fragnet}&  94.2 & 98.5 & 92.8 \\
    FragNet-32 \cite{he2020fragnet}&  95.3 & 98.6 & 96.0 \\
    FragNet-64  \cite{he2020fragnet}&96.3 & 99.1 & 97.6 \\
    \midrule 
    \multicolumn{4}{c}{Self-Supervised}\\
    \midrule
    SWIS \cite{Manna2022SWISSR}& 95.43 & 96.86 & 98.01\\
    Ours &  \textbf{95.58} & \textbf{96.87} & \textbf{98.40} \\
    \bottomrule
    \end{tabular}
    \label {table:6}
\end{table}

\begin{table}[!ht]
    \centering
    \caption{Page level accuracies obtained with Semi-Supervised fine-tuning on 10\% data in Intra-script and Cross-script settings on the Firemaker and CVL datasets.}
    \begin{tabular}{cccc}
    \toprule
        \multirow{1}{*}{Fine-tuning} & \multicolumn{3}{c}{Pre-training Dataset} \\ %\cline{2-4}
        Dataset & IAM & CVL & Firemaker \\
        \midrule
        Firemaker & 78.0 & 76.0 & 77.61 \\
        CVL & 55.40 & 53.42 & 54.37
    \end{tabular}
    \label{table:7}
\end{table}

\subsection{Comparison with Supervised Methods}

In this subsection, we compare the results of the proposed self-supervised framework with the contemporary supervised frameworks in Table \ref{table:5} and Table \ref{table:6}. %It is to be noted that, \checkmark denotes that the methods are supervised in nature, whereas $\times$ denotes self-supervised methods.

\subsection{Discussion}

From the results presented in Table \ref{table:5} and \ref{table:6}, we can see the performance of the proposed model in comparison to the contemporary supervised methods. We can observe that the proposed framework clearly outperforms several supervised methods on the benchmark datasets on word-level and page-level writer identification tasks. The proposed method is also at par in performance with the remaining state-of-the-art supervised methods as well on both word-level and page-level writer identification tasks. Furthermore, the computational complexity of FragNet-16, FragNet-32, and FragNet-64 in terms of FLOPs is 7.14G, 7.41G, and 3.91G, respectively \cite{he2020fragnet}, whereas our model has a computational complexity of about 4 GFLOPs only. Hence, the proposed model achieves a comparable performance, even with lower computational complexity.

\subsection{Semi-Supervised Fine-Tuning}
To test the representation learning capability of the proposed framework, we fine-tuned the pre-trained models in the downstream stage. We used two different types of fine-tuning, 1) Intra-Script, and 2) Cross-Script. In the Intra-script setting, a model pre-trained on a dataset is fine-tuned on 10\% of data from the same dataset. In the cross-script setting, a model pre-trained on a particular dataset is fine-tuned on 10\% of data of a different dataset. In Table \ref{table:7}, we present the results for page-level writer identification obtained in the downstream stage of semi-supervised fine-tuning on the Firemaker and CVL datasets. For these semi-supervised fine-tuning experiments, we use the same configuration described in Sec. \ref{subsec:downstreamexpt}. \\

\section{Statistical Analysis of Disentanglement}
\label{sec:statanal}
% \hl{Review This subsection}

\begin{figure}[!htb]
    \centering
        \begin{tabular}{cccc}
        
           \subfloat{\includegraphics[width = 0.23 \textwidth]{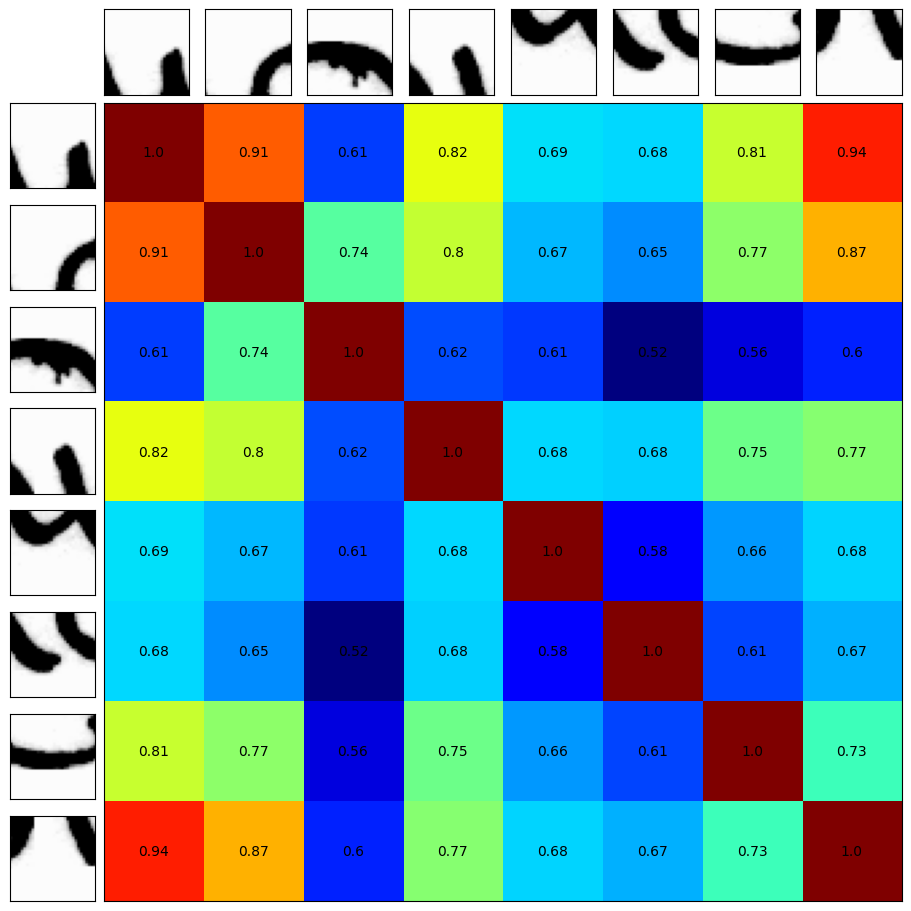}} &
            \subfloat{\includegraphics[width = 0.23\textwidth]{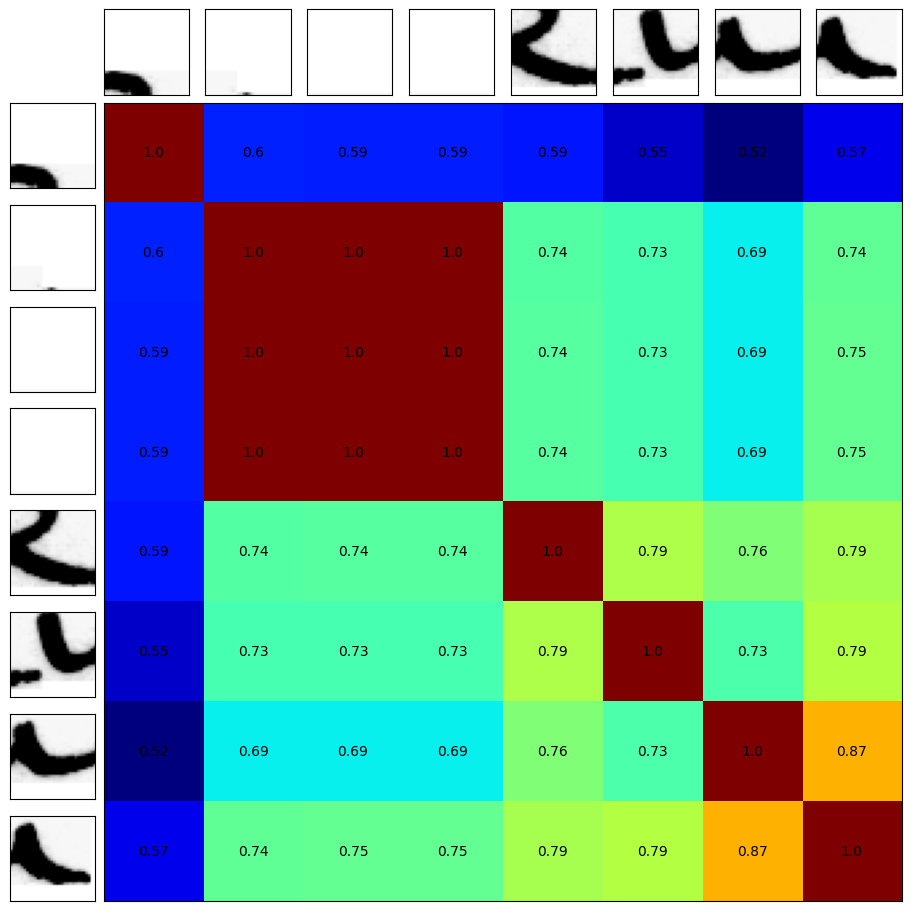}}&
            
        \subfloat{\includegraphics[width = 0.23 \textwidth]{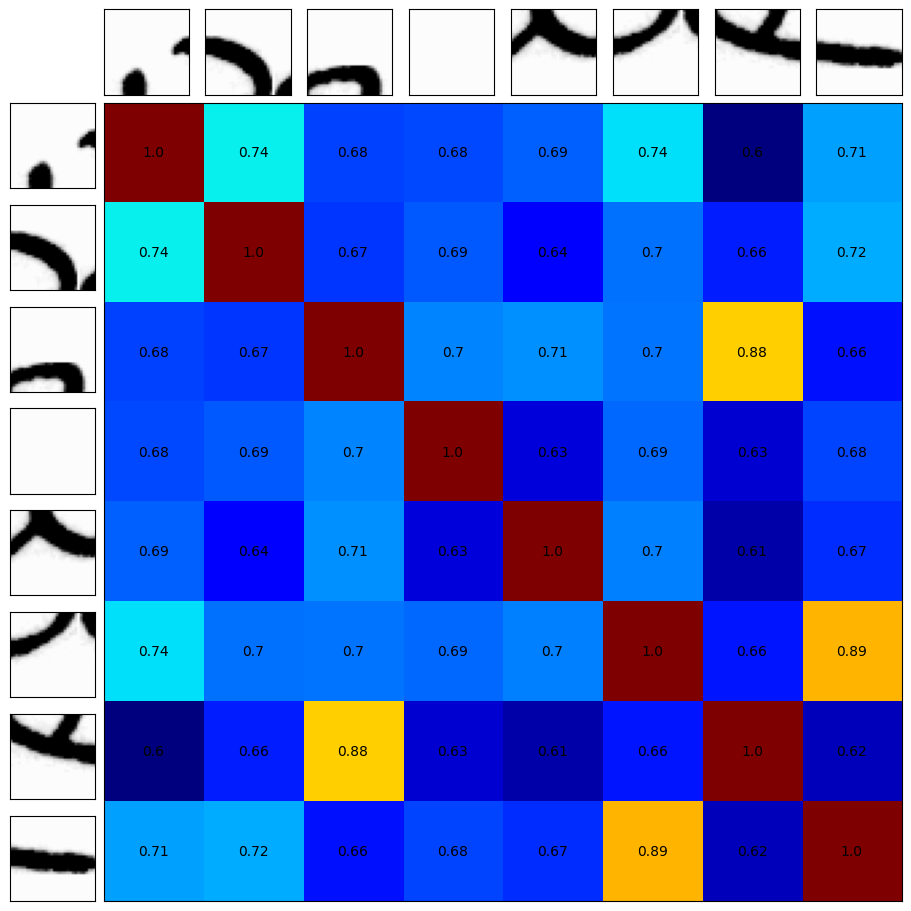}}&
        \subfloat{\includegraphics[width = 0.23 \textwidth]{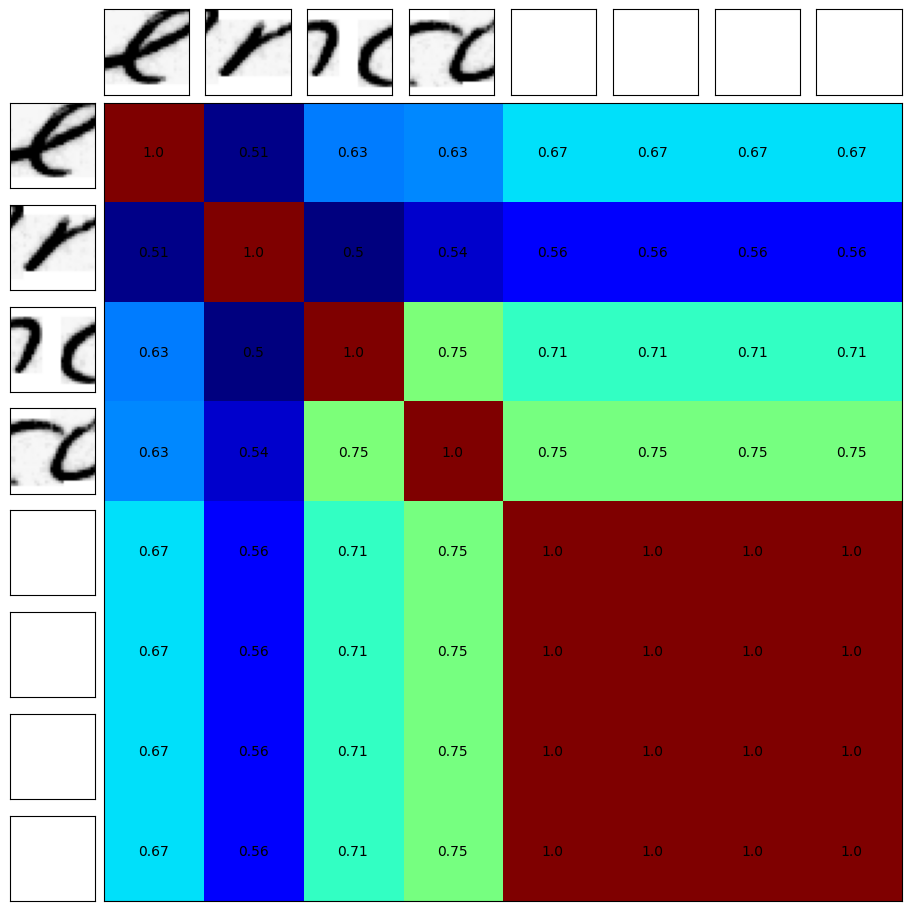}}\\
        \subfloat{\includegraphics[width = 0.23 \textwidth]{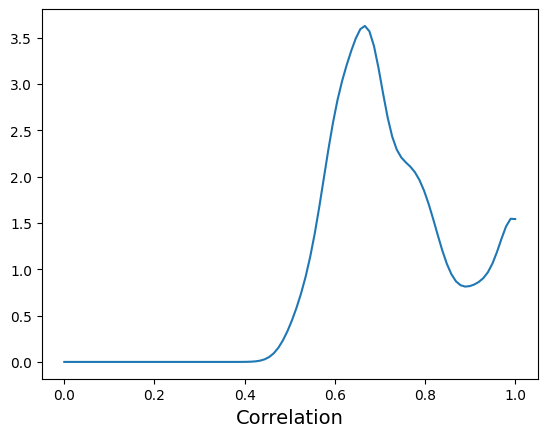}} &
            \subfloat{\includegraphics[width = 0.23\textwidth]{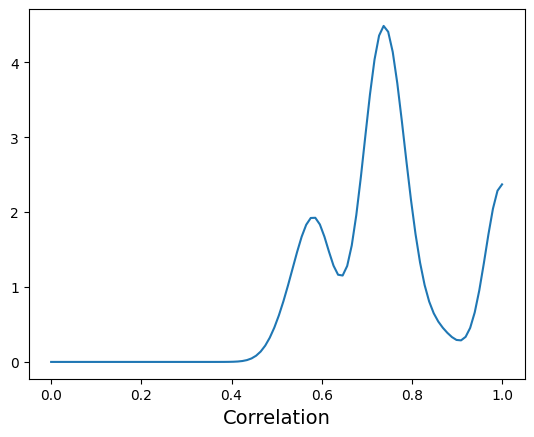}}&
            
        \subfloat{\includegraphics[width = 0.23 \textwidth]{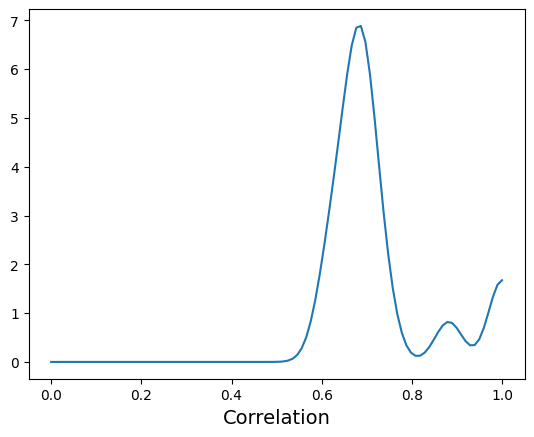}}&
        \subfloat{\includegraphics[width = 0.23 \textwidth]{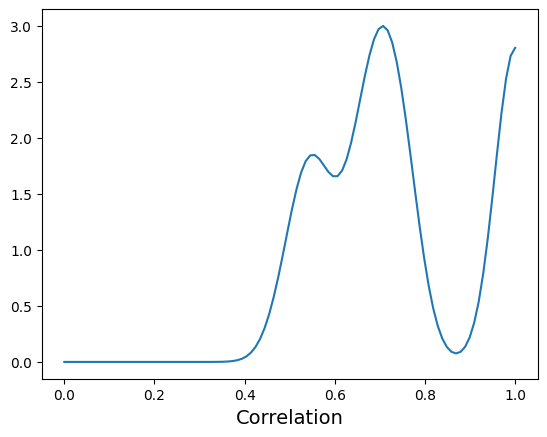}}\\
        \subfloat{\includegraphics[width = 0.23 \textwidth]{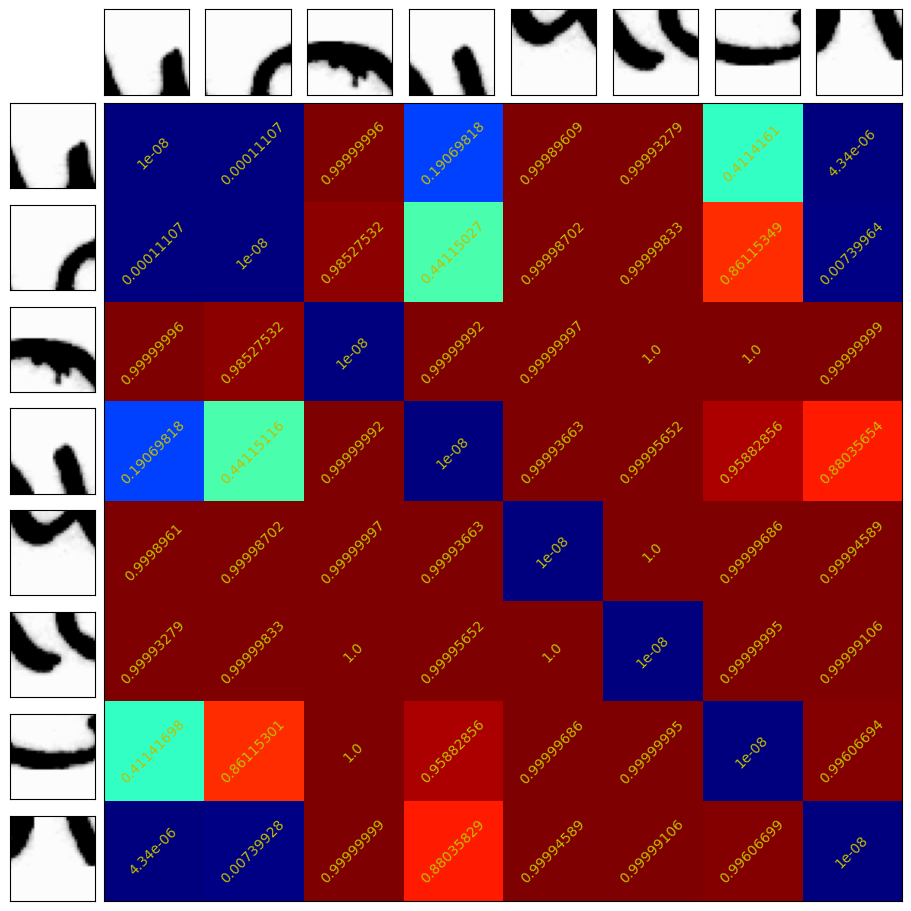}} &
            \subfloat{\includegraphics[width = 0.23\textwidth]{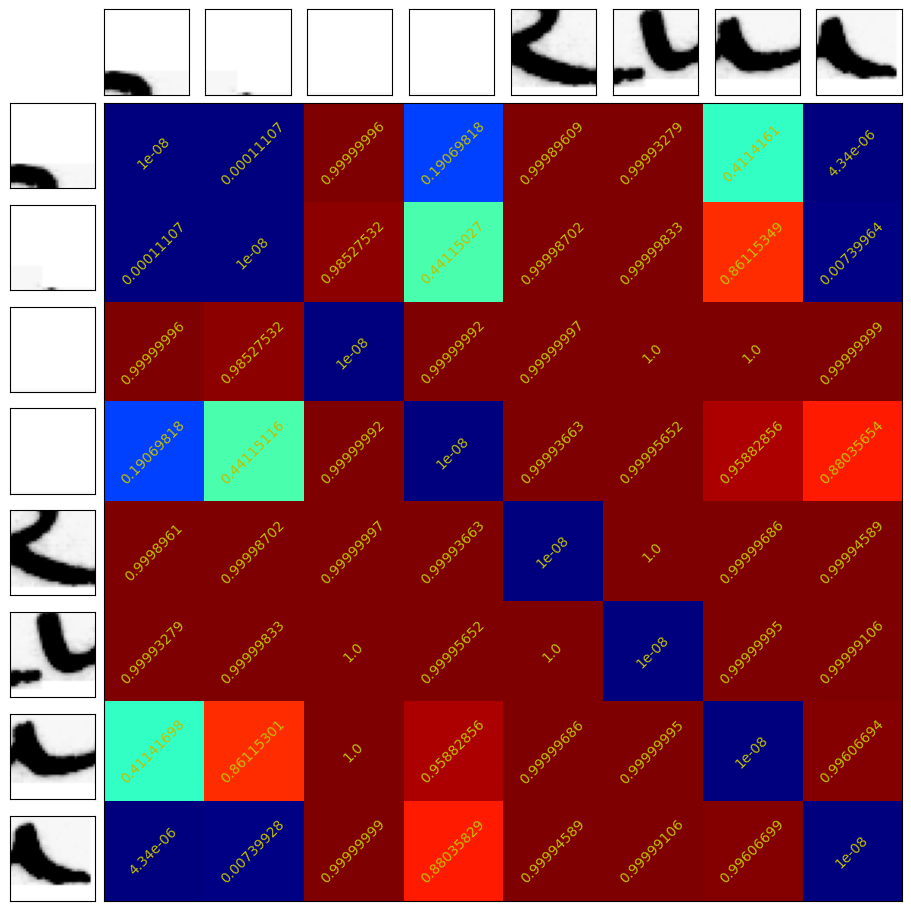}}&
            
        \subfloat{\includegraphics[width = 0.23 \textwidth]{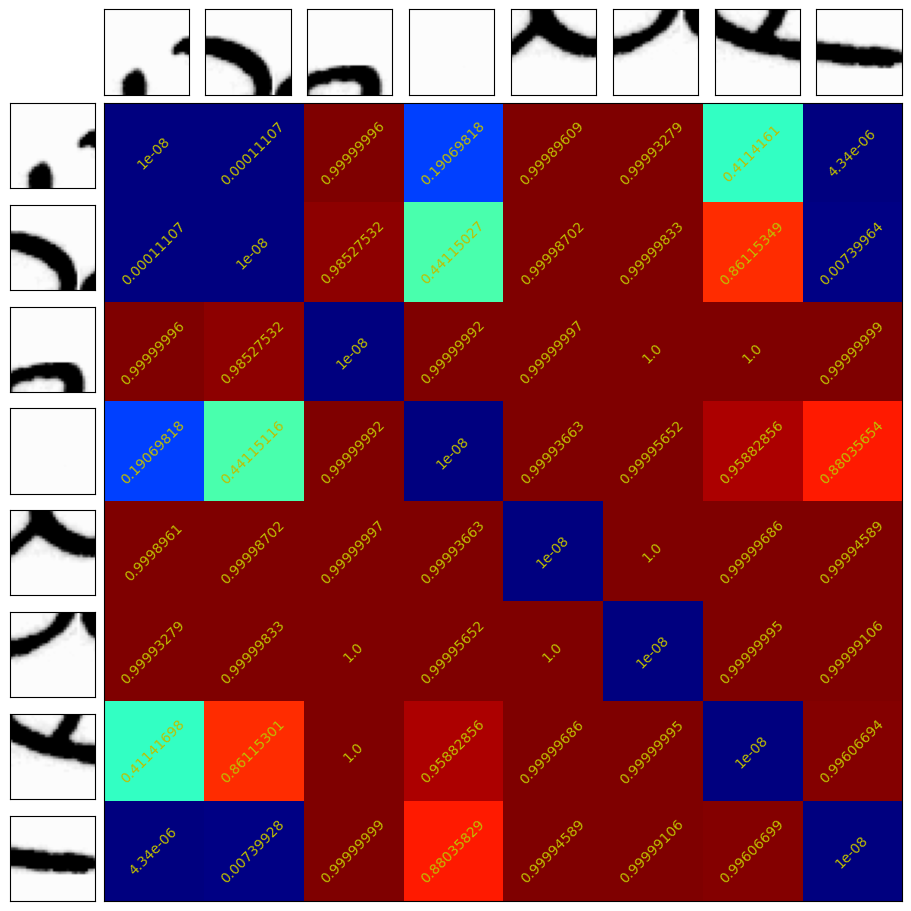}}&
        \subfloat{\includegraphics[width = 0.23 \textwidth]{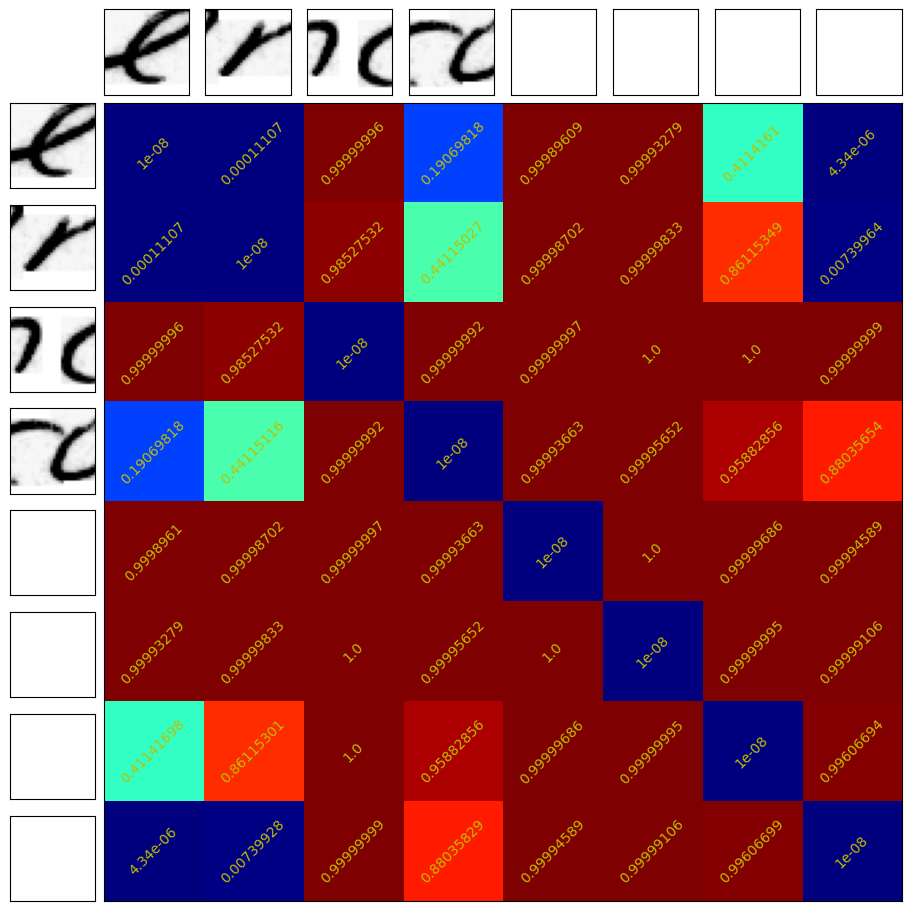}}\\
        \end{tabular}
    \caption{Correlation Map (a-d) of different patches in a single word image with each other. The correlation is calculated with the features extracted from the encoder pre-trained using the proposed framework. The figures (e-h) show the KDE plots of the correlation values. Figures (i-l) show the p-values of the left-tailed t-test conducted on the correlation values as described in Sec. \ref{sec:statanal}.}
    \label{fig:corrmap}
\end{figure}

In this section, we present a qualitative analysis of our pre-training framework. As our objective is to learn disentangled stroke features, we analyze the decorrelation between the different patches we feed as input to the base encoder. However, one thing to note is that as in offline word-level images, we do not have any timestep-wise coordinates of each stroke, the $32 \times 32$ patch cropping does not divide each word into its fundamental strokes by default. To do so, it would require a sophisticated mechanism that would require the incorporation of supervision or annotations, as in online writing data. Having said so, we present the correlation values between the constituent $32 \times 32$ patches.

\begin{figure}
    \centering
    \begin{tabular}{cc}
         \subfloat{\includegraphics[width = 0.5 \linewidth]{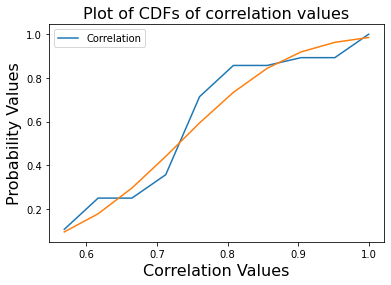}}& 
         \subfloat{\includegraphics[width = 0.51 \linewidth]{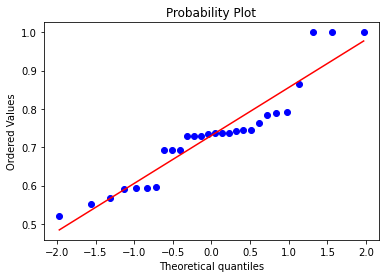}}\\
    \end{tabular}
    \caption{Cumulative Distribution (a) and Quantile-Quantile (Q-Q) Plot (b) for the samples presented in Fig. \ref{fig:corrmap}(a). It shows that the correlation values satisfy the normality assumption and the application of \textit{t-test} is justified.}
    \label{fig:normality}
\end{figure}

We can see that similar images have a high correlation between them, whereas dissimilar images have a lower correlation. As stated earlier, using patches does not explicitly segregate each word into the fundamental strokes, hence we can still see a significant correlation between each patch. We can also see that the correlation between the empty patches with non-empty patches is also low, and the majority of the contribution towards the correlation value comes from the empty regions, as the input data is sparse in nature. From the KDE plots in Fig. \ref{fig:corrmap}.(d-h), we can see a dip in the density estimate around the correlation value of $0.8$. We will use this value of correlation as the threshold for deciding whether two patches are highly correlated or not. We can use a statistical test called $\textit{t-test}$ \cite{CaseBerg:01} to verify our claims. For each image, there are only 28 pairings (discarding the self-pairings and duplicate pairings). Hence, the use of \textit{one-tailed t-test} is justified here. To verify our assumption, we have plotted the CDF and also presented a Q-Q Plot for visual inspection in Fig. \ref{fig:normality} for one of the samples used in Fig. \ref{fig:corrmap}.

Let us frame our hypothesis first,\\
$\mathbf{H_0}$: The correlation between two patches is significant if it is greater than 0.8\\
$\mathbf{H_A}$: The correlation between two patches is not significant if it is less than 0.8

The t-statistic is obtained using the following formula,

\begin{equation}
\begin{split}
    t = \frac{\rho - 0.8}{s.e}, \;\text{where} \; s.e = \sigma/\sqrt{N}
\end{split}
\end{equation}

where, $N = 28$. The degree of freedom is $N - 1 = 27$. We will be using a \textit{left-tailed t-test}.

From the p-values in Fig. \ref{fig:corrmap}(i-l), and considering significance level $\alpha = 0.05$, we can clearly observe that the correlation between two different patches is not significant, except for the patches which are very similar in appearance to each other. Hence, we can reject the null hypothesis for those. The test results hold even under Bonferroni correction \cite{Neyman1928ONTU}.

From this, we can safely claim, that the proposed framework does in fact succeed in learning stroke features such that the correlation between the patches is significantly minimized. 

% \section{Ablation Study on Signature Verification Datasets}

\section{Conclusion}

In this work, we propose a self-supervised feature disentanglement framework for representation learning from handwritten text images. We discuss in detail the role of decorrelation in the prevention of dimensional collapse of representations in self-supervised learning. We then discuss the proposed framework based on the same principle and combine the condition of independence to give us disentangled features. These disentangled features help us to learn stroke features from sparse handwritten text images. We also provide statistical analysis to show that the proposed framework is successful in significantly decorrelating the stroke features by analyzing the features obtained from each patch in a word-level image and validating it through a statistical test. Finally, we also show that the results obtained on the text-independent writer identification tasks in the downstream stage are better or at par with the self-supervised algorithms and the contemporary supervised learning algorithms as well. We can safely claim that this work sets the stage for further work in handwriting analysis research.

%Bibliography
\bibliographystyle{unsrt}  
\bibliography{references}

\end{document}